\documentclass[letterpaper]{article} 
\usepackage{aaai25}  
\usepackage{times}  
\usepackage{helvet}  
\usepackage{courier}  
\usepackage[hyphens]{url}  
\usepackage{graphicx} 
\usepackage{natbib}  
\usepackage{caption} 
\usepackage{algorithm}
\usepackage{algorithmic}

\usepackage{newfloat}
\usepackage{listings}
\DeclareCaptionStyle{ruled}{labelfont=normalfont,labelsep=colon,strut=off} 
\floatstyle{ruled}
\newfloat{listing}{tb}{lst}{}
\floatname{listing}{Listing}
\usepackage{amsmath}
\title{
Language Model Meets Prototypes: Towards Interpretable Text Classification Models through Prototypical Networks}
\author {
   Ximing Wen
}
\affiliations{
    Drexel University, Philadephia, USA\\
   
    xw384@drexel.edu
}

\usepackage{bibentry}

\begin{document}

\maketitle

\begin{abstract}
Pretrained transformer-based Language Models (LMs) are well-known for their ability to achieve significant improvement on NLP tasks, but their \textit{black-box} nature, which leads to a lack of interpretability, has been a major concern. My dissertation focuses on developing intrinsically interpretable models when using LMs as encoders while maintaining their superior performance via prototypical networks. I initiated my research by investigating enhancements in performance for interpretable models of sarcasm detection. My proposed approach focuses on capturing sentiment incongruity to enhance accuracy while offering instance-based explanations for the classification decisions. Later, I develop a novel \textit{white-box} multi-head graph attention-based prototype network designed to explain the decisions of text classification models without sacrificing the accuracy of the original black-box LMs. In addition, I am working on extending the attention-based prototype network with contrastive learning to redesign an interpretable graph neural network, aiming to enhance both the interpretability and performance of the model in document classification.
\end{abstract}

%

\section{Background}

Deep learning models, especially transformer-based Language Models (LMs) have significantly contributed to advancements in Natural Language Processing (NLP), offering encoders as powerful tools for text classification. However, despite their \textit{state-of-art} performance, their complexity and \textit{black-box} nature obscure the decision-making process and hinder their interpretability. Prototype networks, serving as a \textit{white-box} framework where decisions are derived from similarity scores to instance-level prototypes, were initially proposed as an interpretable architecture in the image domain \cite{li2018deep, chen2019looks}. This approach was later adapted to the NLP domain \cite{ming2019interpretable, hong2023protorynet}.

\begin{figure}[t]
  \includegraphics[width=\columnwidth]{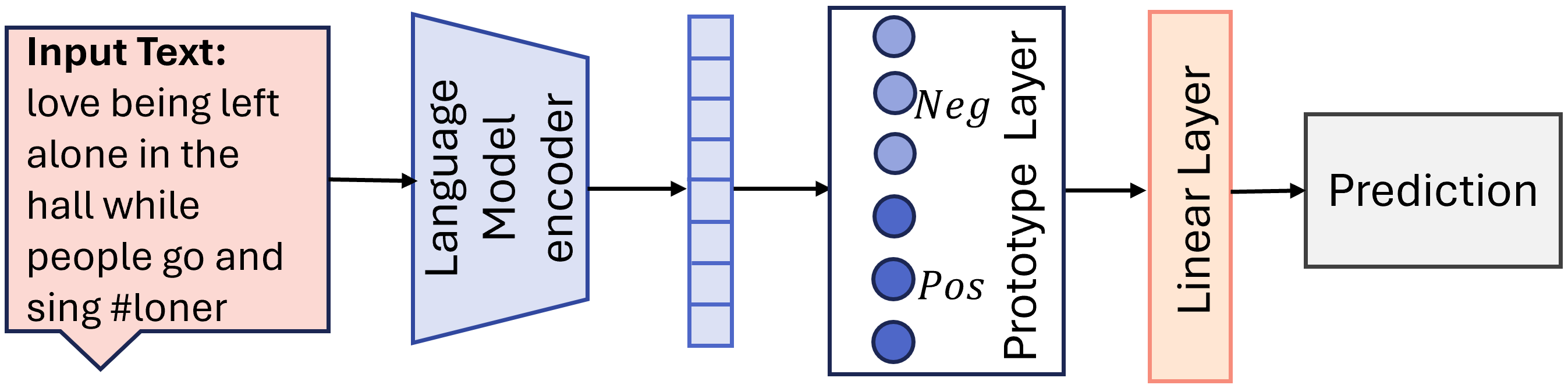} 
  \caption {Illustration of prototype architecture for text classification}
    \label{structure} 
\end{figure}

 Based on the classic framework of prototype learning (\citealp{datta1995learning}), the prototype approach learns prototype vectors through training, projected onto representative cases from previous observations, to explain decisions more intuitively as shown in Figure \ref{structure}. However, existing approaches for text classification still have performance gaps compared to the original \textit{black-box} model. Moreover,  in the context of document classification utilizing Graph Neural Networks \cite{gori2005new}, enhancing the graph prototypes to yield accurate and trustworthy explanations remains a significant challenge.

\section{Goal of the Dissertation}
My dissertation aims to minimize the performance gap of intrinsically interpretable models using prototype network networks in conjunction with LM encoders. I proposed a graph-attention-based prototype network framework to better learn the relatedness (i.e., similar semantic meaning) between the input and prototypes. This framework is applied and experimented with in the context of both sentence-based text classification and graph-based document classification tasks. 
\section{Contributions}

\subsection{A Transformer and Prototype Interpretable Model for Contextual Sarcasm Detection }
My research began with an exploration aimed at minimizing the performance gap in interpretable prototype text classification, focusing initially on the task of sarcasm detection. I proposed an interpretable multi-view framework that integrates semantic and sentiment embeddings from pre-trained LM encoders. Existing explainable methods, including attention-based and post-hoc approaches, generate word-level explanations, which may attribute similar importance to various words in contexts devoid of strong sentiment cues, particularly when sarcasm is conveyed through analogy rather than direct expression. The approach diverges from these methods by leveraging semantic prototypes to provide sentence-level, human-readable explanations.  Furthermore, I innovate with the design of an incongruity loss, which employs sentiment prototypes to discern implicit and explicit sentiments within textual expressions, thereby enhancing predictive accuracy. This methodology achieves \textit{state-of-the-art} performance on benchmark datasets.

\subsubsection{Status:}  Completed. Submitted to TACL.

\subsection{Graph-attention Enhanced Prototype Network for Text Classification}
I develop a novel white-box Multi-head Graph Attention-based prototype framework designed to explain the decisions of text classification models built with LM encoders. \cite{wen2024gaprotonet}.The approach incorporates a prototype layer on top of a fine-tuned LM and utilizes multi-head graph attention \cite{velickovic2017graph} to efficiently learn relatedness by selectively constructing edges between encoded representations and their neighboring prototypes. In the reference time, the decision is solely based on the edge weights computed by each attention head. Different from other prototype networks that use heuristic metrics such as cosine similarity to learn relateness, I leverage graph attention network (GAT), which is known for its ability to capture the importance of neighboring nodes in a graph, enabling more effective learning for each prototype. 
\subsubsection{Performance:}Extensive comparison experiments are conducted with variations of prototype networks on five public benchmark datasets including binary, four-label, and ten-label classification. I also experimented with multiple LMs as encoders. The approach achieved the best preformance compared with all prototype networks on all datasets. Compared with the original \textit{black-box} models, the proposed approach either achieves the best performance, or the performance gap is within 0.3\%.
\subsubsection{Interpretibility:}The case study shows different attention heads could capture different semantic aspects in the input sentence and activate the corresponding prototypes. Moreover, the trained prototype vectors are visualized within the training data space with t-SNE. It is found that the prototype vectors are evenly distributed, indicating that the space formed by these vectors can span over the space so a limited number of prototypes can be used to represent any datapoint. The percentage of distinguished prototypes is between 90\%-95\% when the number of prototypes varies from 10 to 40, suggesting the framework is robust in generating representative prototypes with respect to the change of hyperparameters.

\subsubsection{Status:} Completed. Accepted by COLING 2025.

\subsection{Attention Enhanced Prototype Graph Neural Networks with Contrastive Learning}
My next research plan aims to extend the attention-based prototype network delineated in \cite{wen2024gaprotonet} to the domain of document classification leveraging graph neural networks (GNNs). Contrary to sentence-based text classification, graph-based document classification conceptualizes documents as nodes within a graph, encapsulating relationships such as citations, co-authorships, and keyword associations as edges. The methodology employs transformer-based language models to initiate document node embeddings, while GNNs are utilized to derive comprehensive graph embeddings. Prototype vectors, initially randomly initialized, represent the graph embeddings associated with various classes. Previous endeavors, such as ProtoGNN \cite{zhang2022protgnn}, utilize heuristic distances to measure the similarity between prototypes and input vectors; I hypothesize that the attention-based prototype network harbors the potential to enhance performance in this context.  Moreover, existing approaches use proximity loss to ensure the prototypes can represent real training cases. However, this is more challenging for graph prototypes due to the complexity of graph structure. To address this, I propose the integration of contrastive learning, specifically GraphCL \cite{you2020graph}, to compel prototypes to distill and embody salient graph patterns, thereby augmenting interpretability.

\subsubsection{Status:}\noindent In Progress. Planned to be finished by 04/2024

\bibliography{aaai25}

\end{document}